\pdfoutput=1
\documentclass[11pt,a4paper]{article}
\usepackage[hyperref]{naaclhlt2018}
\usepackage{times}
\usepackage{latexsym}
\usepackage{amsmath}

\usepackage{graphicx}
\DeclareGraphicsExtensions{.pdf,.png,.jpg}

\usepackage{url}
\aclfinalcopy % Uncomment this line for all SemEval submissions

\title{SNU\_IDS at SemEval-2019 Task 3: Addressing Training-Test Class Distribution Mismatch in Conversational Classification}
\author{Sanghwan Bae, Jihun Choi {\normalfont and} Sang-goo Lee \\
  Department of Computer Science and Engineering \\
  Seoul National University, Seoul, Korea \\
  {\tt \{sanghwan,jhchoi,sglee\}@europa.snu.ac.kr} \\}

\date{}

\begin{document}
\maketitle
\begin{abstract}
  We present several techniques to tackle the mismatch in class distributions between training and test data in the Contextual Emotion Detection task of SemEval 2019,
  by extending the existing methods for class imbalance problem.
  Reducing the distance between the distribution of prediction and ground truth, they consistently show positive effects on the performance. Also we propose a novel neural architecture which utilizes representation of overall context as well as of each utterance. The combination of the methods and the models achieved micro F1 score of about 0.766 on the final evaluation. 
\end{abstract}

\section{Introduction}

%%%
A new task whose goal is to predict the emotion of the last speaker given a sequence of text messages has been designed, and coined Contextual Emotion Detection (EmoContext; \citealp{SemEval2019Task3}).
    
Though predicting the emotion of a single utterance or sentence, {\em i.e.} emotion detection, is a well-discussed subject in natural language understanding literature, EmoContext has several novel challenges. Firstly, the class distribution of training data is significantly different from that of the test data. Consequently, a model trained on the training data might not perform well on the test data. There have been efforts made to address the problem of learning from training data sets that have imbalanced class distribution, i.e. the class imbalance problem \cite{chawla2004editorial, buda2018systematic}.  However, they are not applicable to our case, since the imbalance appears only in the test data while the training data can be viewed to be balanced.

We extend the existing methods of addressing the class imbalance problem to be applicable for more general cases where the distributions of the collected training data differ from those of the real population or the data at test time, under the assumption that the validation set is organized carefully to reflect the practical distribution.
From experiments and analyses, we show that the proposed methods reduce the difference between two distributions and as a result improve the performance of the model. 

The additional challenge we have to consider arises from the fact that utterances having identical surface form may have different meanings due to sarcasm, irony, or etc.. Thus a model should track the emotional transitions within a dialogue.
To grasp the context of utterances, we propose a semi-hierarchical encoder structure.
Lastly, the texts contain lots of non-standard words, {\emph e.g.} emoticons and emojis. This makes it difficult to exploit typical pre-trained embeddings such as GloVe \cite{pennington2014glove}.
Therefore, we adopt pre-trained embeddings which are specialized for handling non-standard words. We show that the proposed model largely outperforms the baseline of task organizers by experiments.

%The additional challenge is that the texts contain lots of non-standard words, {\emph e.g.} emoticons and emojis. This makes it difficult to exploit typical pre-trained embeddings such as GloVe \cite{pennington2014glove}. Therefore, we adopt pre-trained embeddings such as emoji2vec \cite{eisner2016emoji2vec} and datastories \cite{baziotis2017datastories}, which are specialized for handling non-standard words and capturing emotional state.

%Lastly, since utterances having the identical surface form may have different meaning due to sarcasm, irony, etc., a model should track the emotional transitions within a dialogue. 
%To grasp the context of utterances, we propose a semi-hierarchical encoder structure. In experiments, this model achieved much higher score than the baseline of task organizers.

\section{Related Work}

\subsection{Contextual Emotion Detection}
EmoContext is a emotion classification data set composed of 3-turn textual conversations.
The goal of the data set is to classify the emotion in the last utterance of each example given the context.
The label set consists of 4 classes: `happy', `sad', `angry' and `others'.
In the training data set, there are about 50\% of `others' class examples and 50\% of emotional (happy, sad, angry) examples, which can be viewed as well-balanced.
On the other hand, only 15\% of examples in the test and the validation data set are labeled as emotional, reflecting the real-life frequency.
For more details, refer to \citet{SemEval2019Task3}.

\subsection{Class Imbalance Problem}
\label{subsec:class-imbalance}
When some classes have the significantly higher number of examples than other classes in a training set, the data is said to have an imbalanced class distribution \cite{buda2018systematic}.
This makes it difficult to learn from the data set using machine learning approaches \cite{batista2004study, mazurowski2008training}, since the learned models can be biased to majority classes easily, which results in poor performance \cite{wang2016training}.
We give a brief explanation of methods to solve this problem.

{\noindent \bf Sampling}:
This type of methods deals with the problem by manipulating the data itself to make the resulting data distribution balanced. The simplest versions are {\em random oversampling} and {\em random undersampling}. The former randomly duplicates examples from the minority classes and the latter randomly removes instances from the majority classes \cite{mollineda2007class}.

{\noindent \bf Thresholding}:
This method moves the decision threshold after training phase, changing the output class probabilities. Typically, this can be done by simply dividing the output probability for each class by its estimated prior probability \cite{richard1991neural, buda2018systematic}. 

{\noindent \bf Cost-Sensitive Learning}:
This assigns different misclassification cost for each class and applies the cost in various ways, {\em e.g.} output of the network, learning rate or loss function.
For multi-class classification tasks, the simplest form of the cost-sensitive learning is to introduce weights to the cross entropy loss \cite{han2017automatic, lin2018focal}:
\begin{equation}
\label{eq:weighted-cross-entropy-loss}
L = -\dfrac{1}{N}\sum_{i=1}^{N}\sum_{c=1}^{K}w^c y_i^c \ln{p(c|x_i)},
\end{equation}
where $N$ and $K$ denote the total number of examples and classes respectively, $p(c|x_i)$ the predicted probability of $i$-th example $x_i$ belonging to class $c$, $y_i^c$ the ground truth label which is 0 or 1, and $w^c$ a class dependent weighting factor.
Recent work suggests to use the inverse ratio of each class, {\em i.e.} $w^c=\frac{N}{N_c}$, where $N_c$ is the number of examples belonging to class $c$ \cite{wu2018weighted, aurelio2019learning}.

{\noindent \bf Ensemble}:
The term ensemble usually refers to a collection of classifiers that are minor variants of the same classifier to boost the performance.
This is also successfully applied to the class imbalance problem \cite{sun2007cost,seiffert2010rusboost},
by replacing the resampling procedure in the bagging algorithm with oversampling or undersampling \cite{galar2012review}.
% Their combination with other techniques to tackle the class imbalance problem have led to several proposals in the literature, with positive results \cite{sun2007cost, seiffert2010rusboost}.
% Simple but still effective approaches are bagging-based ensembles with sampling methods aforementioned \cite{galar2012review}. That is, the resampling procedure in bagging algorithm is replaced with oversampling or undersampling.

\section{Methods for Mismatch Problem}

\subsection{Sampling}
\label{subsec:proposed-sampling}
In our case, it is not possible to make the distinction between majority and minority classes; even if the `others' class is the most prevalent in the training data, the ratio is less than that of the test data.
% random oversampling은 일반적이어서 인용할 논문이 없는듯.
To address this issue, the random minority oversampling technique should be modified,
since it assumes that the class imbalance appears only in the training data set.
Accordingly, we apply random oversampling or random undersampling to make the distribution of the training data similar to that of the validation data.
% For this, the {\bf random oversampling} refers to the replication of randomly selected samples letting the distribution of training data being equal to that of validation data. This means that the examples are only allowed to be duplicated, not discarded. On the contrary, the {\bf random undersampling} is randomly removing the examples to get the same distribution.

\subsection{Thresholding}
\label{subsec:proposed-thresholding}
The basic thresholding method mentioned in \S\ref{subsec:class-imbalance} is not sufficient for our case, since we should additionally \emph{bias} the model output distribution to match the test time distribution, not only correcting the imbalance in the training data.
When $p_r$ is a probability of training time and $p_s$ is that of validation time, we multiply the predicted probability by the estimated class ratio from validation set as below:
\begin{equation}
\label{eq:thresholding}
\begin{split}
y_c(x)=p(c|x) & \approx \frac{\boldsymbol{p_s(c)}}{p_r(c)}\cdot p_r(c|x) \\
& =\frac{\boldsymbol{p_s(c)}}{p_r(c)}\cdot\frac{p_r(c)\cdot p_r(x|c)}{p_r(x)} \\
& =\frac{\boldsymbol{p_s(c)}\cdot p_r(x|c)}{p_r(x)} ,
\end{split}
\end{equation}
where $p(c)=\frac{N_c}{N}$.

\subsection{Cost-Sensitive Learning}
The weighted cross entropy loss, described in Eq. (\ref{eq:weighted-cross-entropy-loss}), can be used for our case, as long as the weights are carefully chosen.
Although the reciprocal of class ratio is helpful for learning balanced prediction \cite{wu2018weighted}, the target distribution between classes is not uniform for our task.
Also, the misclassification cost of `others' class should be larger than those of emotional classes, because the model tends to predict it less than the actual.
Therefore, it is reasonable to modify $w^c$ by multiplying the estimated ratio of each class for test time, {\em i.e.} $w^c=\frac{N^r}{N_c^r}\cdot \frac{N_c^s}{N^s}$,
where $N^r$ and $N^s$ denote the number of instances in training and validation data set respectively. This corresponds to the term introduced in Eq. (\ref{eq:thresholding}), as $\frac{N^r}{N_c^r}\cdot \frac{N_c^s}{N^s}=\frac{p_s(c)}{p_r(c)}$.
The difference between them is that thresholding utilizes the term in inference phase after training is finished, while weighted cross entropy loss includes it from the training time. 

\subsection{Ensemble}
We combined bagging-based ensemble with sampling techniques represented in \S\ref{subsec:proposed-sampling}. In addition, we compared ensembles of randomly initialized classifiers using other methods ({\em i.e.} thresholding and cost-sensitive learning).

\section{Model and Training Details\footnote{The implementation of our model is available at \url{https://github.com/baaesh/semeval19_task3}}}

\begin{figure}[t]
\begin{center}
    \includegraphics[width=0.9\columnwidth]{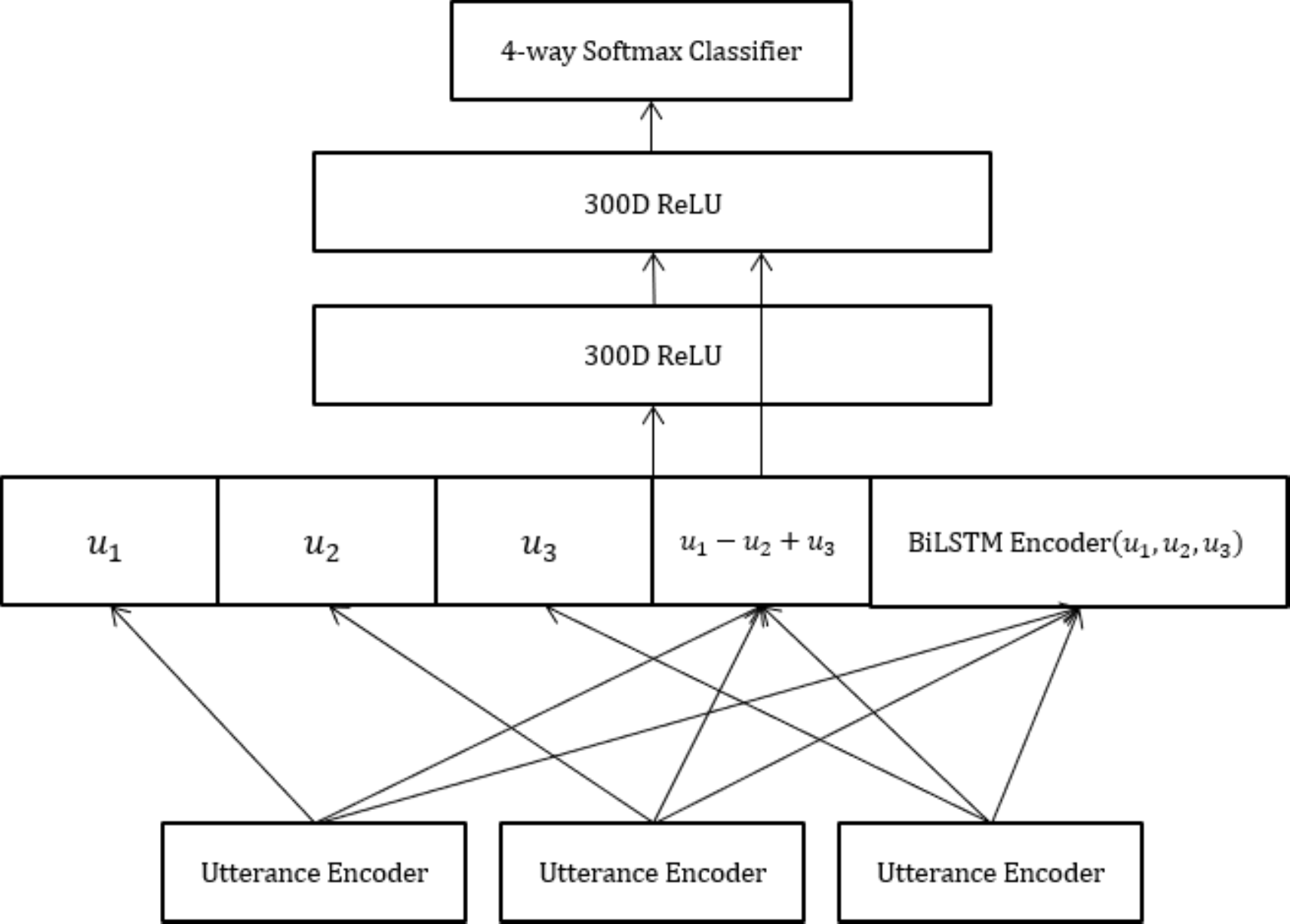}
    \caption{Overall Architecture} \label{fig_alchitecture}
    \label{fig:overall-architecture}
\end{center}
\end{figure}

\subsection{Overall Architecture}
\begin{figure}[t]
\begin{center}
    \includegraphics[width=0.85\columnwidth]{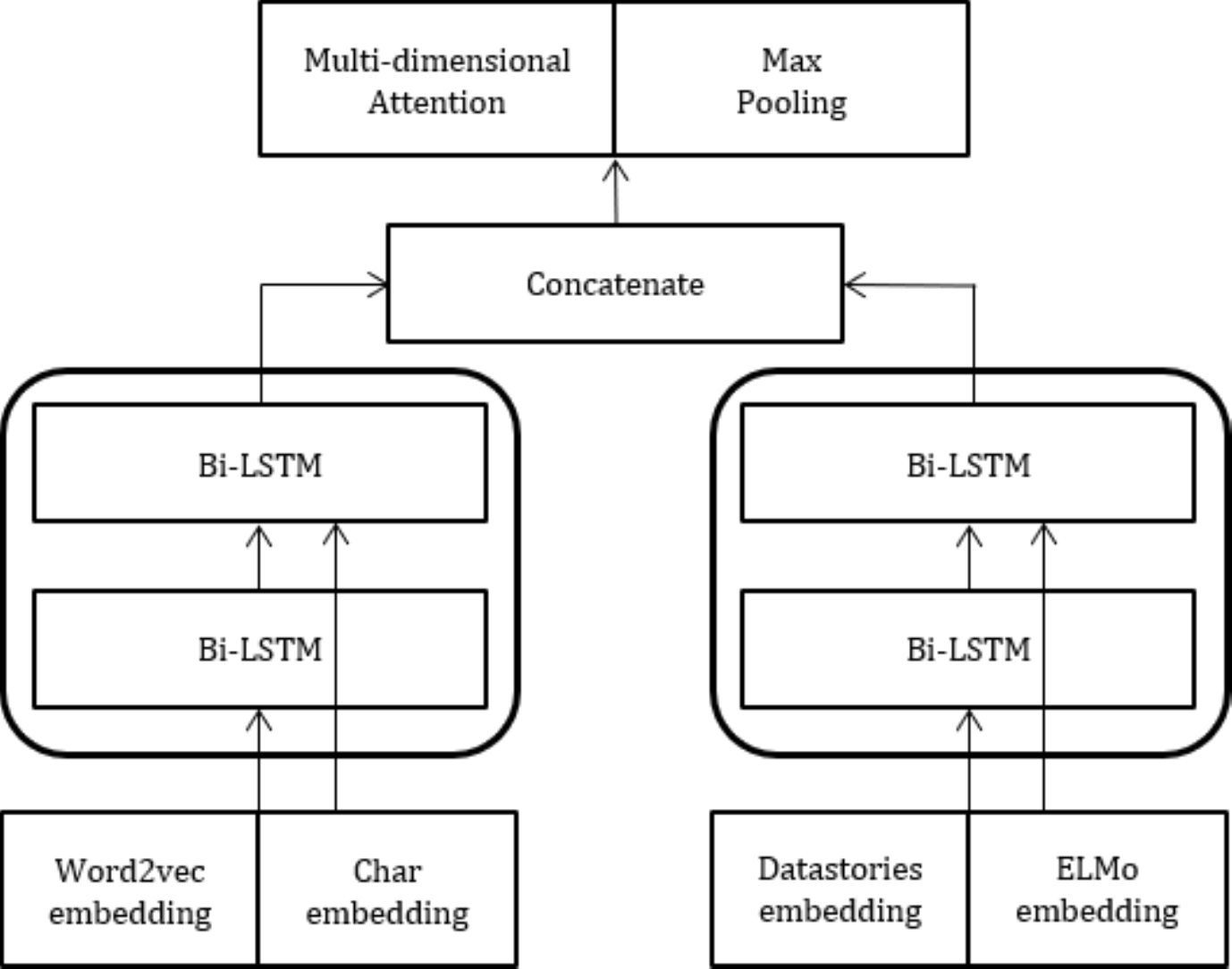}
    \caption{Utterance Encoder} \label{fig:utterance-encoder}
\end{center}
\end{figure}

We propose a semi-hierarchical structure to capture the context as well as the meaning of each single utterance.
Fig. \ref{fig:overall-architecture} depicts the overall architecture. Each single utterance representation $u_m$ is encoded by Utterance Encoder described in Fig. \ref{fig:utterance-encoder}. In addition, we introduce another bi-directional LSTM (BiLSTM) encoder for higher level representation which receives the outputs of utterance encoder as its inputs.
The representation of all context is generated by the concatenation of $u_1$, $u_2$, $u_3$, $u_1-u_2+u_3$ and the output of this encoder.
Then it is fed to the two 300-dimensional (300D) hidden layers with $ReLU$ activation with shortcut connections and a $softmax$ output layer. 

\subsection{Utterance Encoder}
The proposed utterance encoder has two types of shortcut-stacked bi-directional long short-term memory (BiLSTM) encoder \cite{nie2017shortcut} to fully exploit four types of lexicon level representations.
The first encoder utilizes pre-trained word2vec \cite{mikolov2013distributed} embeddings concatenated with trainable embeddings from a character level convolutional neural network (CNN).
We also added emoji2vec \cite{eisner2016emoji2vec} embeddings to word2vec to map emoji symbols to the same space.
The other encoder receives concatenated representations of pre-trained datastories \cite{baziotis2017datastories} embeddings and contextualized represenations from ELMo \cite{peters2018deep}. 

The results from the two stacked BiLSTM encoder are concatenated.
We use the multi-dimensional source2token self-attention of \citet{shen2018disan} and max pooling to integrate contextualized word level representations to a single utterance representation, as in \citet{im2017distance}.

\subsection{Training Details}
We use 300D Google News word2vec\footnote{\url{https://code.google.com/archive/p/word2vec/}} embeddings and 300D pre-trained emoji2vec.\footnote{\url{https://github.com/uclmr/emoji2vec}}
Datastories vectors\footnote{\url{https://github.com/cbaziotis/datastories-semeval2017-task4}} which were pre-trained on a big collection of Twitter messages using GloVe are also 300D.
The dimension of character embeddings is fixed to 15, and it is fed to a CNN where the filter sizes are 1, 3, and 5 and the number of feature map for each is 100, thus a 300D vector is generated for each word as a result.
To guarantee the same size for ELMo embeddings, a 300D position-wise feed-forward network is applied above them.
The hidden states of all the BiLSTMs for each direction are 150D and the number of layers is 2.

Our model is trained using the Adam optimizer \cite{kingma2014adam} with a learning rate of 0.001 and a batch size 64.
We clip the norm of gradients to make it smaller than 3. 
Dropout \cite{srivastava2014dropout} technique is applied to word embeddings with $p=0.1$.
We chose the best model based on a micro F1 score on the validation set.

\label{sect:pdf}

\section{Experiments}

\begin{figure}[t]
\begin{center}
    \includegraphics[width=0.9\columnwidth]{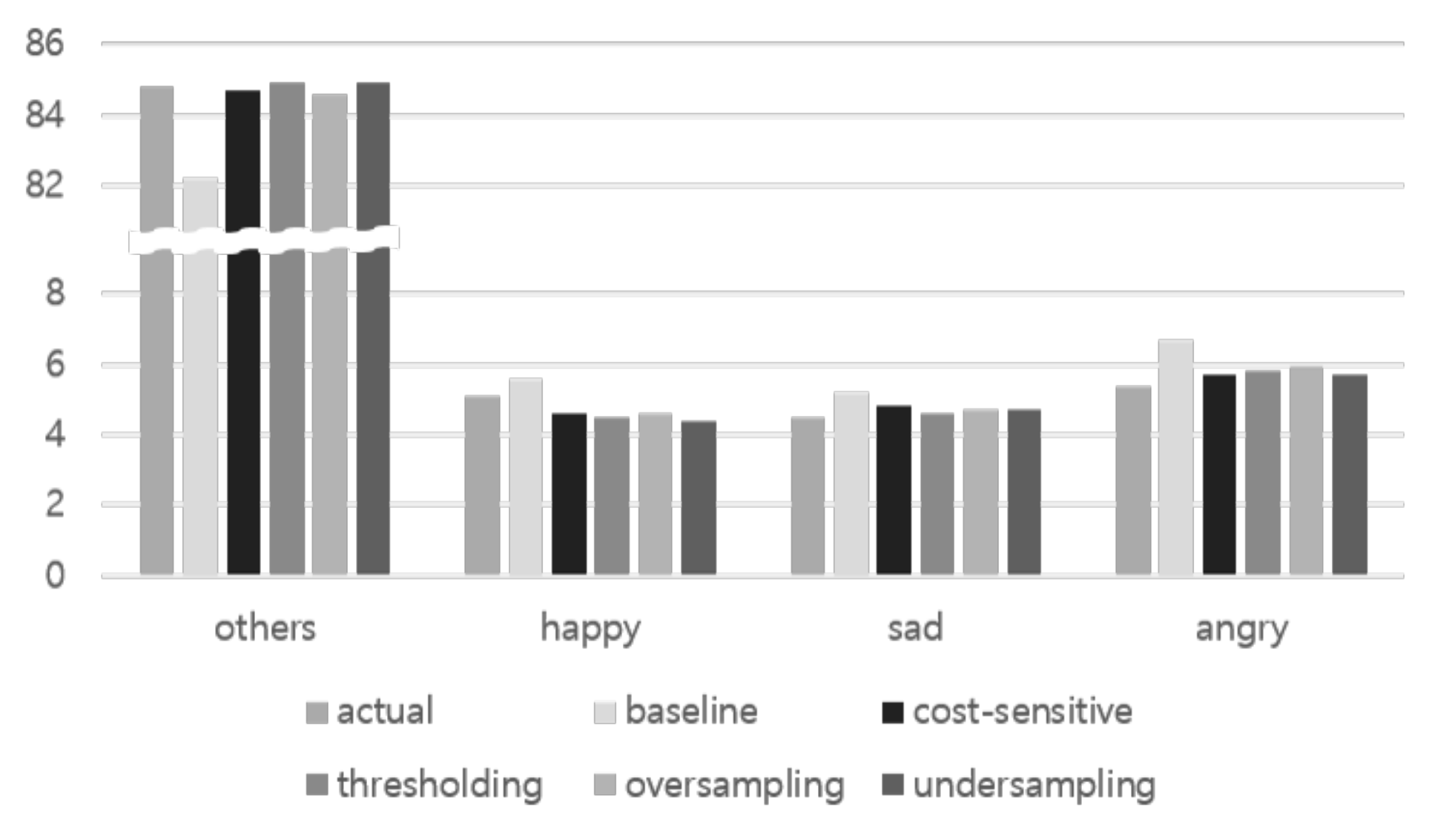}
    \caption{Class distributions on test data. Actual is from the ground truth labels and the others are from the predicted labels by each model. These are the averages of 10 results with random initialization.} \label{fig:class-distribution}
\end{center}
\end{figure}

\subsection{Effects of mismatch in class distributions}
%여기는 내가 제대로 이해한지 모르겠어서 다시봐야될듯
Fig. \ref{fig:class-distribution} shows that the class distribution of predictions from the baseline model without any methods applied is substantially different from that of the actual test data.
On the other hand, when the proposed methods are applied, the gap becomes much smaller.
From this result, we conjecture that the difference in output distributions could be a reason for poor performance of the baseline compared to the proposed methods, as presented in Table \ref{table:single-model-comparison}.

\subsection{Single model methods}
Table \ref{table:single-model-comparison} shows the accuracy and micro F1 scores of variants of our methods and baselines on the test set. We report the mean and standard deviation of 10 experimental runs (with the same hyperparameters) for each methods. And all the models were chosen based on their performance on the validation set.

The reported results show that proposed methods except undersampling are effective for enhancing both accuracy and F1 score.
This means that alleviating the difference of class distribution is the key factor for the higher performance.
In the case of undersampling, since the total size of training data decreases, the model seems to fail to capture the general semantics.
The result is consistent with \citet{buda2018systematic}, where the undersampling solely does not bring a performance gain for deep learning models.
On the other hand, in our experiments, thresholding and cost-sensitive learning were the most effective approaches when a single model is used.

\begin{table}[t!]
\small
\centering
\begin{tabular}{lcccc}
  {\bf Approach} & {\bf Acc} & $(\pm)$ & {\bf F1} & $(\pm)$\\
  \hline \hline
  Baseline (organizers) & - & - & .587 & - \\
  Baseline (ours) & .914 & .005 & .726 & .008 \\
  \hline
  Oversampling & .922 & .004 & .733 & .012 \\
  Undersampling & .919 & .006 & .719 & .013 \\
  Thresholding & {\bf .924} & .002 & .738 & .010 \\
  Cost-Sensitive & .924 & .004 & {\bf .739} & .010 \\
  \hline \hline
\end{tabular}
\caption{Comparison of single model approaches on the test set. 
  }
\label{table:single-model-comparison}
\end{table}

\begin{table}[t!]
\small
\centering
\begin{tabular}{lcc}
  {\bf Approach} & {\bf Acc} & {\bf F1} \\
  \hline \hline
  Baseline (ours) & .921 & .743 \\
  \hline
  Oversampling & .930 & {\bf .758} \\
  Undersampling & .930 & .753 \\
  Thresholding & .930 & .752 \\
  Cost-Sensitive & {\bf .931} & .757 \\
  \hline
  Mixed (submitted) & {\bf .933} & {\bf .766} \\
  \hline \hline
\end{tabular}
\caption{Comparison of ensemble approaches on the test set. 10 models were used for each ensemble result.
  }
\label{table:ensemble-comparison}
\end{table}

\subsection{Ensemble methods}
Table \ref{table:ensemble-comparison} reports the comparison of ensemble models.
The results show that our methods consistently outperform the baseline.
We can see that ensemble with bagging has a great effect on refining class distribution,
and in this time, undersampling also showed a good performance.
Overall, for ensemble methods, oversampling and cost-sensitive learning performed best.
The version we submitted to the leaderboard was the ensemble of different methods selected by their performance on validation set, and achieved the official result of 0.766.

\section{Conclusion}
In this paper, we proposed several methods for alleviating the problems caused by difference in class distributions between training data and test data.
We demonstrated that these methods have positive effects on the result performance.
We also presented a novel semi-hierarchical neural architecture that effectively exploits the context and the utterance representation.
For future work, we plan to conduct more systematic experiments on other data sets to generalize our results.

% \section*{Acknowledgments}

% The acknowledgments should go immediately before the references.  Do
% not number the acknowledgments section. Do not include this section
% when submitting your paper for review. \\

%\bibliography{semeval2018}
%\bibliographystyle{acl_natbib}

\end{document}